%% file: main.tex
\pdfoutput=1

\documentclass[11pt]{article}

\usepackage[final]{acl}

\PassOptionsToPackage{sorted}{natbib}
\input{preamble}

\title{ADQA: Evaluating Appreciation and Understanding in Audio Descriptions}
\title{\textit{What You See is What You Ask}: Evaluating Audio Descriptions}

\author{Divy Kala, \hspace{2mm} Eshika Khandelwal,\hspace{2mm} Makarand Tapaswi \\
CVIT, IIIT Hyderabad, India \\
{\normalsize\url{https://katha-ai.github.io/projects/adqa/}}}

\begin{document}

\maketitle

\input{sections/0_abstract}

\input{sections/1_intro}

\input{sections/2_relwork}

\input{sections/3_twosources}

\input{sections/4_benchmark}

\input{sections/6_experiments}

\input{sections/7_conclusion}
\input{sections/8_limitations}

\bibliography{longstrings, main}

\clearpage
\appendix

\input{appendix/alignment}

\input{appendix/benchmark}
\input{appendix/qualitative}

\input{appendix/userstudy}

\input{appendix/prompts}

\end{document}

%% file: preamble.tex
\usepackage{times}
\usepackage{latexsym}

\usepackage[T1]{fontenc}

\usepackage[utf8]{inputenc}

\usepackage{microtype}

\usepackage{inconsolata}

\usepackage{graphicx}

\usepackage{amsmath}
\usepackage{amssymb}
\usepackage{mathrsfs}
\usepackage{mathtools}
\usepackage{booktabs}
\usepackage{tabularx}
\usepackage{rotating}
\usepackage{multirow}
\usepackage{booktabs}
\usepackage{lipsum}
\usepackage{enumitem}
\usepackage{pifont}
\usepackage{soul}
\usepackage{mdframed}
\usepackage{siunitx}

\usepackage[accsupp]{axessibility}
\usepackage{amsfonts}       
\usepackage{nicefrac}       
\usepackage{microtype}      
\usepackage[dvipsnames, table]{xcolor}
\usepackage{url}            
\usepackage{hyperref}

\newcommand{\benchmark}{ADQA}

\newcommand{\xmark}{\ding{55}} 
\newcommand{\cmark}{\ding{51}} 

\renewcommand{\paragraph}[1]{\vspace{1mm}\noindent\textbf{#1}}
\renewcommand{\subparagraph}[1]{\vspace{1mm}\textbf{#1}}

\newcommand{\meanstd}[2]{#1\footnotesize{\color{gray} $\pm$#2}}

\usepackage{xspace}
\makeatletter
\DeclareRobustCommand\onedot{\futurelet\@let@token\@onedot}
\def\@onedot{\ifx\@let@token.\else.\null\fi\xspace}

\def\eg{\emph{e.g}\onedot} 
\def\ie{\emph{i.e}\onedot} 
 
 \def\vs{\emph{vs}\onedot}

\makeatother

\usepackage[capitalize]{cleveref}
\crefname{figure}{Fig.}{Figs.}
\Crefname{figure}{Figure}{Figures}
\crefname{section}{Sec.}{Secs.}
\Crefname{section}{Section}{Sections}
\crefname{table}{Tab.}{Tabs.}
\Crefname{table}{Table}{Tables}
\crefname{appendix}{App.}{Apps.}
\Crefname{appendix}{Appendix}{Appendices}

\usepackage[most]{tcolorbox}
\tcbset{
  findingbox/.style={
    colback=gray!10,
    colframe=black,
    boxrule=0.3pt,
    arc=1pt,
    left=2pt,
    right=2pt,
    top=2pt,
    bottom=2pt,
    boxsep=2pt
  }
}

\renewenvironment{quote}{%
  \list{}{%
    \leftmargin0.5cm   
    \rightmargin\leftmargin
  }
  \item\relax
}
{\endlist}

%% file: sections/0_abstract.tex
\begin{abstract}

Audio descriptions (ADs) narrate important visual details in movies, enabling Blind and Low Vision (BLV) users to \textit{understand narratives} and \textit{appreciate visual details}.
Existing works in automatic AD generation mostly focus on few-second trimmed clips, and evaluate them by comparing against a single ground-truth reference AD.
However, writing ADs is inherently subjective.
Through alignment and analysis of two independent AD tracks for the same movies, we quantify the subjectivity in
\textit{when} and \textit{whether} to describe, and
\textit{what} and \textit{how} to highlight.
Thus, we show that working with trimmed clips is inadequate.
We propose \textit{ADQA}, a QA benchmark that
evaluates ADs at the level of few-minute long, coherent video segments, testing whether they would help BLV users \textit{understand the story} and \textit{appreciate visual details}.
ADQA features visual appreciation (VA) questions about visual facts and narrative understanding (NU) questions based on the plot.
Through ADQA, we show that current AD generation methods lag far behind human-authored ADs.
We conclude with several recommendations for future work and introduce a public leaderboard for benchmarking.

\end{abstract}

%% file: sections/1_intro.tex
\section{Introduction}
\label{sec:intro}

\begin{quote}
\textit{“It's just that my eyes don't work. My brain is perfectly intact. Let me think for myself.”}
\vspace{-3mm}
\begin{flushright}
-- anonymous AD consumer
\end{flushright}
\vspace{-2mm}
\end{quote}

Blind and Low Vision (BLV) individuals watch movies and TV shows with assistance from audio descriptions (ADs). Writing ADs is a complex task that requires experts to identify the most relevant visual elements and describe them in a coherent and concise manner to fit within the gap between dialogs~\cite{pavel2020rescribe}.
A seminal book on this topic, The Visual Made Verbal~\cite{visual_made_verbal}, states the oft-referenced “first rule of description” as
"What You See Is What You Say."
It also emphasizes two central goals of AD: enabling BLV users to
(i)~\textit{appreciate} the visual richness of a scene; and
(ii)~\textit{understand} the narrative by providing important (visual) plot points.

In addition to these broad goals, several guidelines (\eg~\cite{youdescribe_playlist}) provide concrete recommendations:
(i)~focus on visual elements, avoid audible content (unless audio source is ambiguous \eg~mixer grinder),
(ii)~stick to visual facts and do not opine or interpret (\eg~beautiful woman),
(iii)~read on-screen text when relevant to the story (\eg~two years later),
(iv)~provide information just in time, neither too early nor late,
(v)~balance the amount of information and speed of narration, and
(vi)~match vocabulary to the material and be concise.

To make content more accessible, there is a rising interest in automatic AD generation~\cite{han2023autoad, han2024autoad3, xie2024autoad0, gao2024adreview, chu2024llmad, park2025narrad, wang2025uniad}.

\begin{figure*}[t]
\centering
\includegraphics[width=0.98\linewidth]{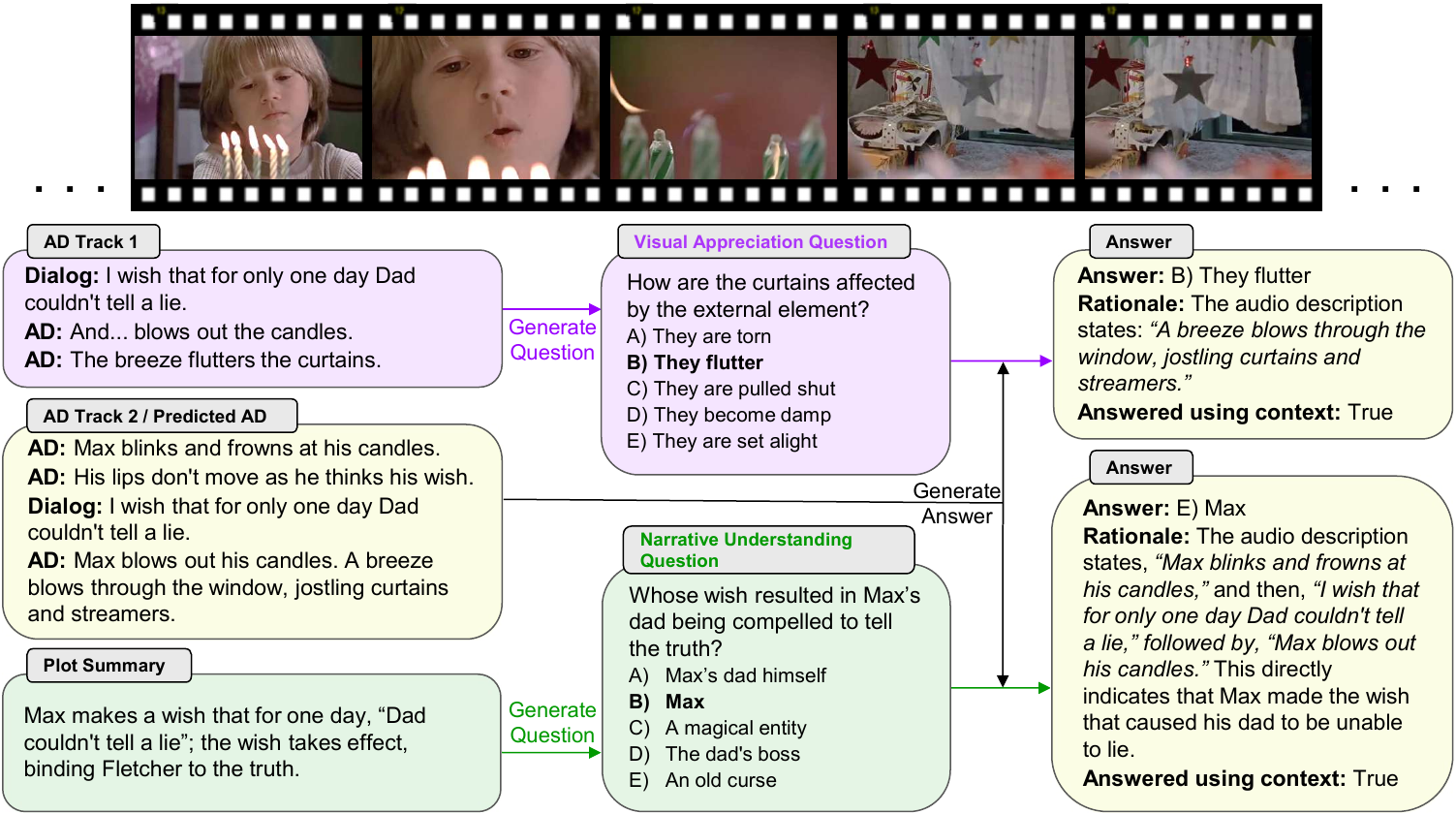}
\caption{
We present ADQA's question generation and answering framework.
A small part of a video from the film \textit{Liar Liar} from CMD-AD~\cite{han2024autoad3} is shown.
AD Track 1 and 2 from AudioVault show the dialogs and ADs describing the video in different ways. The plot summary is taken from CMD~\cite{bain2020condensedmovies}.
AD Track 1 is used to create \textcolor{RoyalPurple}{\textbf{Visual Appreciation}} questions, whereas the plot summary is used to create \textcolor{ForestGreen}{\textbf{Narrative Understanding}} questions. LLMs are prompted to answer both question types using the AD track under evaluation, here, AD Track 2.
The video can be watched here:
{\small \url{https://youtu.be/IsBB4i4k2PM}}.
}
\label{fig:teaser}
\end{figure*}

However, even after development of several new methods and metrics (\cref{sec:relwork}), evaluation treats ADs as isolated captions. 
Typically, each predicted AD is compared to a single ground-truth reference for each \textbf{trimmed clip}---few second video clips trimmed to the duration where the ground-truth ADs are spoken.
A challenge with this evaluation is the subjective nature of ADs, \ie~two experts may write different ADs for the same video (\cref{fig:teaser}).
For a subset of movies that have two independently narrated AD tracks, we align them and quantify how this subjectivity manifests in \cref{sec:twosources}.

Working with trimmed clips also does not check whether generated ADs would help BLV users appreciate the visual richness of the media or better understand the unfolding story.

To address these gaps, we posit that AD evaluations need to be performed for longer video segments and check whether they assist \textit{visual appreciation} and \textit{narrative understanding}.
We propose the Audio Description Question Answering (\benchmark{}) benchmark (\cref{fig:teaser}), a multiple-choice QA framework (\cref{sec:adqa}) that evaluates ADs of few-minute long videos.
\textbf{ADQA evaluates generated ADs by mixing them with dialog and using them as context for an LLM to answer questions.}
In fact, barring sound effects and speaker identity, dialog and AD are the primary modalities for BLV users.

We create two kinds of QAs.
For visual appreciation, we create questions based on ground-truth ADs that ask about specific visual details of a scene.
Answering these requires AD models to pick the most prominent visual story element and describe it correctly.
For narrative understanding, we create questions based on the story plot.
Answering these requires generating ADs that advance the storyline together with dialog.

In summary, our contributions are:
(i)~ADQA, a new benchmark to evaluate AD generation methods by focusing on a central tenet: would the generated ADs help BLV users \textit{appreciate} and \textit{understand} the story?;
(ii)~a first detailed analysis of the subjective nature of ADs;
(iii)~comparative evaluation of AD generation methods through ADQA, hosted as a public leaderboard; and
(iv)~concrete recommendations for future of AD generation.

%% file: sections/2_relwork.tex
\section{Related Work}
\label{sec:relwork}
A comprehensive review of AD generation methods, including advances in VLMs/LLMs, is provided by~\citet{gao2024adreview}.
Below we discuss related datasets, methods, and metrics.

\paragraph{AD datasets.}
LSMDC~\cite{rohrbach2017lsmdc} contains professionally written ADs manually aligned with full-length movies to create trimmed clips.
MAD~\cite{soldan2022mad} uses AudioVault%
\footnote{AudioVault is a non-profit repository of movie ADs. See \url{https://audiovault.net}.}
for training and LSMDC for evaluation.
CMD-AD~\cite{han2024autoad3} also uses AudioVault and pairs ADs with videos from the CMD dataset~\cite{bain2020condensedmovies},
while TV-AD~\cite{xie2024autoad0} features ADs for TV shows.
ADQA, our benchmark, leverages the 2 most commonly used datasets: MAD-eval and CMD-AD.

\paragraph{Generation methods}
are of two types: fine-tuned and zero-shot.
\textul{Fine-tuned approaches} adapt pretrained models with light-weight modules and add character identity~\cite{han2023autoad, han2023autoad2, han2024autoad3, wang2025uniad, ye2025focusedad, ye2024mmad}.

Among them, DistinctAD~\cite{fang2025distinctad} penalizes repeated content to reduce redundancy and DanteAD~\cite{deganutti2025dantead} improves long-term context using transformers.
Fill-in~\cite{park2020fillin} and MiCAP~\cite{raajesh2024micap} address identity grounding without explicitly naming characters and building consistent identity clusters across clips.
Among the \textul{zero-shot methods}
AutoAD-Zero~\cite{xie2024autoad0} and LLM-AD~\cite{chu2024llmad} prompt vision-language models (VLMs) with frames and character cues,
NarrAD~\cite{park2025narrad} uses scripts, and MMNarrator~\cite{zhang2024mmnarrator} proposes a memory-augmentation to handle long video context, 
Shot-by-Shot~\cite{xie2025shotbyshot} uses neighboring shots, film grammar, and VLMs to improve AD generation.
We evaluate some of these models on ADQA.

\paragraph{AD evaluation.}
Since writing ADs is a subjective process, a large variety of metrics have been proposed over the years.
Classically, ADs are evaluated using \textul{CIDEr}~\cite{rohrbach2017lsmdc, soldan2022mad}

which compares n-grams between the generated and reference ADs.  
However, CIDEr penalizes linguistic diversity when only a single reference AD is available, the typical case for AD evaluation~\cite{vedantam2015cider}.

AutoAD II introduces an alternative to one-to-one AD matching, \textul{R@k/N}, which rewards semantic relevance.
AutoAD III proposes \textul{CRITIC} for character names and \textul{LLM-AD-Eval} for sentence-pair semantic scoring using LLMs.
Shot-by-Shot~\cite{xie2025shotbyshot} introduces the \textul{Action Score} to evaluate action coverage.

MMNarrator~\cite{zhang2024mmnarrator} proposes \textul{SegEval} to
assess both textual properties (originality, consistency) and sequence-level attributes (coherence, diversity, specificity).
However, like reference-based methods, it assumes that system-generated ADs and reference ADs highlight the same visual details, which is not always true (see \cref{sec:twosources}).

NarrAD~\cite{park2025narrad} introduces \textul{human-centered metrics} like usefulness and recommendability via user studies, which, while highly effective, are not scalable.

Different from above, our work focuses on evaluating ADs through a QA framework to check whether they would help BLV users appreciate and understand the story.

Finally, CinePile~\cite{rawal2024cinepile} uses ADs to generate QAs for long-form video comprehension.
We differ in multiple ways:
our aim is to evaluate generated ADs,
appreciation questions often stem from a single AD (trimmed clips), and
narrative questions are created based on plots.

%% file: sections/3_twosources.tex
\section{Takeaways from Two AD Tracks}
\label{sec:twosources}

While MAD-train, CMD-AD, and TV-AD have relied on AudioVault
as the primary source of ADs, they all use a single AD track.

\subsection{Aligning Multiple AD Tracks}
\label{subsec:twosources:alignment}

For a subset of movies in CMD-AD,
we identify and analyze multiple AD tracks from AudioVault.
Some are US \vs~UK movie variants%
\footnote{In our analysis, the content across these variants is quite similar, allowing us to use them for our work.}
while others are multiple tracks for the same movie.

AudioVault hosts complete movie audio tracks comprising dialog, AD, music, and sound effects.
To align AD tracks, we follow three steps:
(i)~obtain timestamped transcriptions using WhisperX~\cite{bain2023whisperx};
(ii)~classify each transcription sentence as AD or dialog using LLMs (\cref{app:prompts}); and
(iii)~align the two transcriptions using dynamic time warping~\cite{han2024autoad3} anchored via dialog that have strong matches.
Apart from a few movies with missing scenes that are treated manually, the above process yields good alignments.

While this procedure is similar in spirit to AutoAD-3~\cite{han2024autoad3}, there are two important differences.
(i)~We observe that using an LLM results in much better AD/dialog classification than identifying the narrator's voice.
(ii)~We align two transcribed tracks containing dialog + AD,
while AutoAD-3 aligns transcribed CMD videos (dialog) with AudioVault transcriptions (dialog + AD).

Our alignment process provides a timeline of sequentially aligned dialog from both tracks, interspersed with non-aligned ADs.
We use this dialog alignment to compute a linear transformation (slope and offset) that maps the timestamps from track 1 to track 2 and apply the transformation.

We then create a mapping between ADs from the two tracks if their durations have an overlap score (computed in \cref{app:alignment}, step 4a) higher than a chosen threshold (default: 50\%).
Any ADs for which we cannot find a mapping are marked as \textit{non-aligned}.
Details of the full alignment procedure, including additional mapping steps, are provided in \cref{app:alignment}.

\begin{figure}[t]
\centering
\includegraphics[width=1\linewidth]{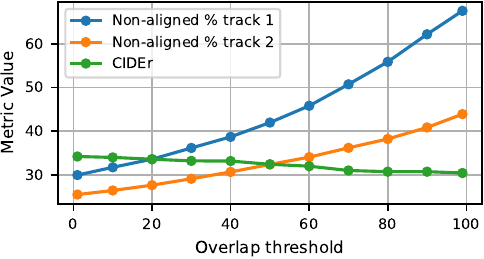}
\vspace{-6mm}
\caption{Impact of overlap threshold on AD alignment on the two-track subset of CMD-AD movies.
The fraction (\%) of non-aligned ADs increases with threshold (expected).
Interestingly, even at low thresholds, 25-30\% ADs are not aligned indicating that many ADs in one track are not present in the other.
Additionally, CIDEr does not increase with better temporal overlap (high threshold) suggesting that even well-aligned ADs often differ substantially in wording.
}
\vspace{-2mm}
\label{fig:align_threshold}
\end{figure}

\cref{fig:align_threshold} shows that increasing the overlap threshold increases the number of non-aligned AD pairs, as expected.
Yet, even at a very low threshold (1\%), about 25–30\% of ADs still remain \textit{non-aligned}.
This suggests that some ADs inherently lack counterparts in the other track, fundamentally challenging the validity of sentence-pair evaluation.

\subsection{Similarity between Aligned ADs}
\label{subsec:twosources:similarity}

\input{tables/two-source-align}

For a subset of 17 CMD-AD movies with two AD tracks, \cref{tab:two-source-align} shows the \% of aligned ADs,
the average overlap \%,
and the average similarity scores between the aligned ADs:
BERT cosine similarity~\cite{devlin2019bert} and CIDEr~\cite{vedantam2015cider}.
Even for aligned ADs with high overlap, we observe poor CIDEr scores highlighting the challenges of using n-gram based metrics.

Next, \cref{fig:align_threshold} shows that CIDEr stays constant with increasing overlap threshold, and even reduces a bit at higher thresholds.
This indicates that even with very high temporal overlap, aligned AD pairs may still use different words.

To further investigate the nature of aligned AD pairs, \cref{fig:bert_vs_cider} compares their BERT similarity against CIDEr.
We see five important scenarios corresponding to low/high values of the similarity metrics. 

They highlight the subjective nature of ADs: where aligned ADs can describe different details (33.8\%, quadrant III),
or the same detail, but using different words resulting in poor CIDEr scores (16.3\%, quadrant IV).
Finally, about 27\% pairs fail catastrophically with 0 CIDEr.

We summarize our \textit{takeaways}:
(i)~Different AD experts may subjectively choose \textit{when} to describe an event or \textit{whether} to describe it at all, resulting in a large proportion of non-aligned ADs.
(ii)~Even among ADs aligned in time, experts may differ in \textit{what} visual details to highlight for a coherent story, resulting in low semantic similarity scores.
(iii)~Finally, experts may differ in \textit{how} they phrase the same detail, resulting in low scores from n-gram-based metrics like CIDEr, especially when evaluated against a single reference.

Overall, these findings suggest that evaluating ADs in a strict one-to-one manner, as in video captioning, is unsuitable, since experts may disagree at multiple levels.
We need a better evaluation aimed at the heart of AD creation---do they help BLV users appreciate and understand the story.

\input{tables/bert-vs-cider}

%% file: tables/two-source-align.tex
\begin{table}[t]
\centering
\small
\tabcolsep=0.12cm
\begin{tabular}{cccc}
\toprule
Aligned \%  & Overlap \% & BERT similarity & CIDEr \\
\meanstd{60.7}{16.7} &
\meanstd{85.6}{16.8} &
\meanstd{85.3}{6.4} &
\meanstd{37.3}{97.6} \\
\bottomrule
\end{tabular}
\vspace{-2mm}
\caption{Results of aligning and mapping ADs between two sources for 17 movies of the CMD-AD dataset.
CIDEr (37.3) is low when comparing the aligned ADs (60.7\%) that have a high average overlap (85.6\%).
}
\vspace{-2mm}
\label{tab:two-source-align}
\end{table}

%% file: tables/bert-vs-cider.tex
\begin{figure}[t]
\centering
\tabcolsep=0.12cm
\small
\includegraphics[width=1\linewidth]{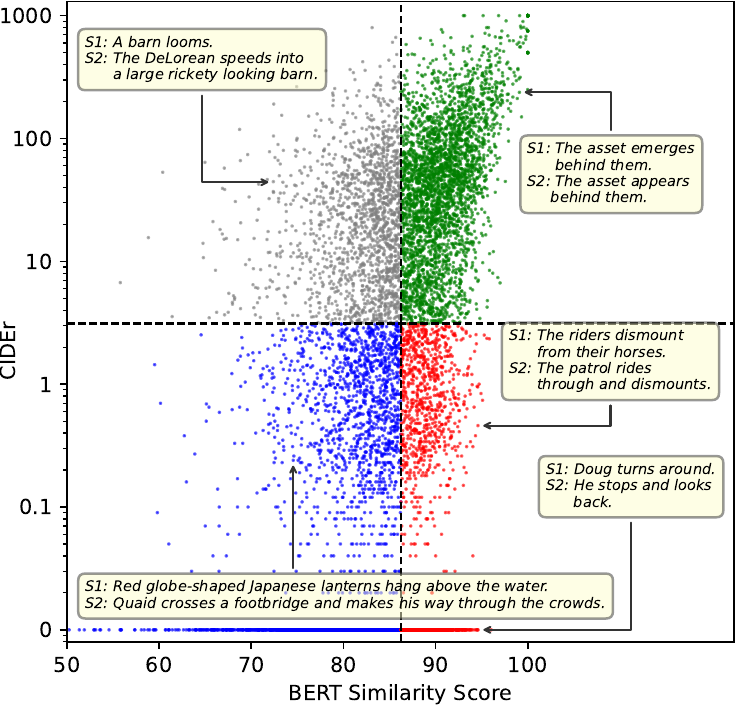} \\
\vspace{1mm}
\begin{tabular}{c c c l}
\toprule
B            & C            & P \% & \multicolumn{1}{c}{Reason / Implication} \\
\midrule
$\uparrow$   & $\uparrow$   & 33.8 &
Same detail with similar words \\
$\downarrow$ & $\uparrow$   & 16.1 &
Matched words but poor semantic match \\
$\downarrow$ & $\downarrow$ & 33.8 &
Different details for the same clip \\
$\uparrow$   & $\downarrow$ & 16.3 &
Same detail with different words \\
-            & 0            & 26.6 &
Catastrophic failure, different words \\
\bottomrule
\end{tabular}
\vspace{-1mm}
\caption{BERT Similarity (B) \vs~CIDEr (C) for \textit{time-aligned ADs} from two AD tracks on 17 movies from the CMD-AD dataset.
The quadrants and $\uparrow$ or $\downarrow$ labels are separated by median scores (B: 86.2, C: 3.1) and the proportion of samples in each quadrant is in P~\%.
We summarize the reasons for these scores in the table.
}
\vspace{-2mm}
\label{fig:bert_vs_cider}
\end{figure}

%% file: sections/4_benchmark.tex
\section{\benchmark{} Benchmark for Evaluating ADs}
\label{sec:adqa}

Our goal is to evaluate ADs for their ability to help BLV users appreciate and understand the story.
Given the subjective nature of the task and challenges of sentence-pair metrics (\cref{sec:twosources}), we move from few-second long trimmed clips to a \textbf{video segment}---a few minute interval from the movie showcasing a coherent story segment (\eg~the 2-3 minute Youtube videos used in CMD-AD) (\cref{subsec:adqa:clips_to_video}).
Next, we pose AD evaluation as a QA task.
We measure the quality of predicted ADs (and thereby the generation method) based on our ability to correctly answer questions using those ADs.
\cref{subsec:adqa:creating} presents the multiple-choice QA (MCQA) creation strategy and \cref{subsec:adqa:features} the quantitative details of \benchmark{}.

\subsection{From Trimmed Clips to Video Segments}
\label{subsec:adqa:clips_to_video}

Traditionally, AD datasets are built by extracting trimmed clips corresponding to the duration where AD narration occurs, using transcription timestamps.
However, AD narrations aren’t strictly synchronized with the visual events being described; narration can anticipate or lag behind the visuals.
While LSMDC~\cite{rohrbach2017lsmdc} adjusts the timestamps manually to ensure that the input videos contain the visual details, recent works such as MAD-train or CMD-AD move away from this tedious process.
As a result, generated ADs often score poorly in 1-to-1 comparisons on trimmed clips, since they may lack the visual details described in the ground-truth ADs.
Instead, we consider longer video segments (typically 2-3 minutes, which may include multiple ADs) as the evaluation unit. 
This shift not only provides more context for generating ADs, but also encourages producing coherent, non-redundant descriptions rather than isolated captions.
For brevity, we will refer to trimmed clips as \textit{clips} and longer video segments with potentially multiple ADs as \textit{videos}.

CMD-AD is a special case as the videos are obtained from a YouTube channel that shares famous or important plot moments of a movie.
These videos are \SI{140}{\second} on average and can be directly treated as video segments.
The videos are also accompanied by plot-like descriptions (used in text-to-video retrieval by~\citet{bain2020condensedmovies}) that are appropriate for creating narrative understanding questions.

On the other hand, MAD provides ADs for the entire movie. 
Inspired by work on aligning plots and movies~\cite{tapaswi2015plotalign}, we use LLMs to align plot synopses sentences with dialog + AD transcriptions (referred as \textit{script}) of the entire movie, thereby identifying \textit{scene} boundaries.
In the prompt to the LLM (see~\cref{app:prompts}), we observe the following clauses to be especially important:
(i)~merge consecutive \textit{script} scenes when they align to long plot sentences  describing multiple events;
(ii)~ensure that details in the plot are present in or can be inferred from the script; and
(iii)~not require every scene to be aligned to some plot sentence.
This creates video segments of about \SI{116}{\second} that are accompanied by plot sentences.
The first part of \cref{tab:adqa-stats} presents some statistics for CMD-AD and MAD-eval, highlighting the number of video segments and those with an aligned plot sentence.

\subsection{Creating QAs}
\label{subsec:adqa:creating}

We prompt Gemini-2.5-Pro to generate all questions in ADQA.
Similar to video QA benchmarks~\cite{tapaswi2016movieqa, lei2018tvqa},
our questions have 1 correct answer among 5 options.

\paragraph{Visual appreciation MCQAs.}
We obtain appreciation QAs by prompting the LLM to come up with questions based on the factual details conveyed in ADs (typically a single AD).
While the input context includes dialog + ADs of a video segment, we prompt the model to
(i)~not create questions based on dialog; and
(ii)~not have questions that are answerable by looking at other questions (or answer options) of the same video.
Along with question generation, the LLM is prompted to create multiple plausible answer options.
We notice that an advanced model such as Gemini-2.5-Pro performs much better on the question generation task than smaller/faster models such as Gemini-2.0-Flash.
An example provided in the prompt is:
Given the AD: "A green truck speeds through the highway crossing a yellow barrier",
multiple questions can be created:
"What vehicle is seen on the highway?",
"What is the color of the vehicle on the highway?", or
"What color is the barrier on the highway?".

\paragraph{Narrative understanding MCQAs.}
Next, we create narrative understanding questions based on the plot descriptions accompanying the video segment.
As indicated before, CMD-AD videos are accompanied by a plot-like description, while MAD videos are aligned with plot synopsis sentences.
As these descriptions present the narrative, creating factual questions based on them results in narrative understanding MCQAs.
We discourage generation of ambiguous questions that are not highlighted in the plot.
In the prompt, we present the following example:
For a plot description: "The shining spaceship lands on a strange planet covered in glowing blue plants and mist.",
questions such as
"Where does the spaceship land?" (A: on a strange planet),
or
"What makes the planet unusual?" (A: the planet is covered in glowing blue plants and mist)
are considered good.
On the other hand, questions such as
"What did the spaceship do?" (it could have done several different things apart from landing)
are marked ambiguous.

\paragraph{Answer rationale during question generation.}
As seen in the question generation prompts (\cref{app:prompts}), beyond the question and answer options, the LLM is prompted to also select the correct answer and provide a rationale for the same.
The rationale for appreciation QAs starts with
"As specified in the AD, ...", indicating that the model should point to the specific AD(s) used to create the question.
Similarly, the rationale for narrative understanding QAs is encouraged to refer to plot-like description.
Beyond MCQA generation, asking the LLM to pick the correct answer and provide a rationale acts as a self-verification check resulting in higher quality QAs.
We also rely on rationales during AD evaluation, as explained in the next section.

In summary, for each video segment, we generate:
(i)~Visual Appreciation (VA) question(s);
(ii)~Narrative Understanding (NU) question(s);
(iii)~correct answer and wrong options for each question; and
(iv)~a rationale for the correct answer referring to the dialog + AD script or plot.

\input{tables/adqa-stats}

\subsection{ADQA Analysis}
\label{subsec:adqa:features}

\cref{tab:adqa-stats} shares some numbers of the ADQA benchmark.
For CMD-AD, we present numbers on the full \textit{evaluation} set (98 movies) used to evaluate AD generation approaches.
As we obtain ADs from two tracks for 17 movies, we also report numbers for this \textit{TwoTrack subset}.
For MAD-eval, we obtain two AD tracks for all 10 movies: the original LSMDC ADs and different ADs sourced from AudioVault.

ADQA features multiple appreciation questions per AD ($\sim$2.4) and several understanding questions for each video segment ($\sim$5.5).
While some example QAs from our benchmark are presented in \cref{fig:teaser},
several more questions, answer options, and answer rationales are shared in \cref{app:qualitative:adqa}.
We make \textit{public} the questions, answers, and correct choices for 5 movies from CMD-AD and 1 movie from MAD-eval.
Questions from the remaining movies are kept \textit{private} for evaluation on the online leaderboard.
More details are presented in \cref{app:benchmark}.

%% file: tables/adqa-stats.tex
\begin{table}[t]
\centering
\small
\tabcolsep=0.11cm
\begin{tabular}{l cc c}
\toprule
& \multicolumn{2}{c}{CMD-AD} & MAD-eval \\
                            & Full  & TwoTrack  & Full  \\ 
\midrule
\#Movies                    & 98    & 17    & 10    \\ 
\#VideoSegments (VS)        & 591   & 112   & 551   \\ 
\#VideoSegments w/ plot     & 591   & 112   & 338   \\ 
Duration (s)                & 140   & 142   & 116   \\ 
\#ADs                       & 7316  & 1484  & 6331  \\
\midrule
\#VisualAppreciation Q      & 17595 & 2705  & 15441 \\ 
\#Vis App Q per AD          & 2.41  & 1.82  & 2.44  \\ 
\midrule
\#NarrativeUnderstanding Q  & 3128  & 585   & 1962  \\ 
\#Nar Und Q per VS         & 5.29  & 5.22  & 5.80  \\ 
\bottomrule
\end{tabular}
\vspace{-2mm}
\caption{ADQA benchmark in numbers.}
\vspace{-2mm}
\label{tab:adqa-stats}
\end{table}

%% file: sections/6_experiments.tex
\section{Experiments}
\label{sec:experiments}

We present the evaluation setup, followed by a user study assessing LLM reliability, and end with a comparison of various AD generation methods.

\subsection{Answering Setup and Metrics}
\label{subsec:exp:evalsetup}

\paragraph{Base setup.}
We use a common prompt and LLM (Gemini 2.0 Flash) to perform all evaluations in our work.
The setup involves providing all questions (and corresponding answer options) of a video segment, followed by a specified context.
The LLM is prompted to answer each question independently.
The key idea is to evaluate ADs generated using different methods by feeding them (separately) as context to the LLM.

\paragraph{Rationale-based answering.}
We find that LLMs trained on world knowledge know details of popular movies.
This means that they are able to answer questions from prior knowledge or common sense rather than using the provided context.
To evaluate whether the LLM answers based on the provided context, we generate three outputs:
(i)~the predicted answer among 5 choices;
(ii)~a descriptive rationale for choosing the answer; and
(iii)~a binary label indicating whether the rationale suggests that the answer was derived from context.

For completeness, we report three values:
(i)~CA: Correct Answer;
(ii)~AC: Answer uses Context (\ie~the binary label is \texttt{True}); and
(iii)~CC: Correct answer \textit{and} uses Context.
A prediction is considered correct only if the chosen answer option is correct \textit{and} the binary label confirms context-based reasoning.
Thus, CC is our primary metric of interest and can be considered as ADQA's \textit{accuracy}.

\paragraph{Types of context.}
We evaluate five different types of context to understand how different information affects accuracy.
They are:
(i)~no context,
(ii)~only movie name,
(iii)~only dialog,
(iv)~only ADs, and
(v)~dialog + ADs.
These ablations allow us to quantify the contribution of each context type to the accuracy.

\paragraph{Accuracy ratio.}
As studied in \cref{sec:twosources}, AD creation is inherently subjective and even human-generated ADs may not fully agree with each other.
We therefore do not expect any method to score 100\% accuracy on ADQA.
Instead, we define accuracy ratio to measure how much of the gap between human performance and only dialog is closed by a model:
\begin{equation}
\text{Accuracy Ratio}_m = 
\frac{\text{CC}_m - \text{CC}_{\text{dialog}}}
{\text{CC}_h - \text{CC}_{\text{dialog}}} \, ,
\label{eq:accratio}
\end{equation}
where $\text{CC}_m$ corresponds to accuracy of method $m$ (the method under evaluation),
$\text{CC}_{\text{dialog}}$ is the accuracy using only dialog context,
and $\text{CC}_h$ is computed by running the same LLM answering pipeline, but using human-authored AD tracks (e.g., from AudioVault) instead.
Here, CC$_h$ acts as top-line \textit{human performance} for AD generation.

\subsection{Assessing LLM Reliability }
\label{subsec:exp:userstudy}
\input{tables/user-study}
To address potential unreliability in automatically generated questions and answers, we conducted a human evaluation with 8 participants covering both question generation and answering stages. The participants were unpaid volunteers from within our research group, with age ranging from 20-35 years. The study did not gather personally identifiable information. Volunteers were notified that research publications would present only the average scores calculated from their ratings.

We sampled 32 CMD-AD videos (16 each for Visual Appreciation and Narrative Understanding), yielding 128 questions correctly answered (CC) using track~2’s AD and dialog.
Depending on the task, users were shown: (i)~the question generation context; (ii)~the generated question and options; (iii)~the answering context; and (iv)~the chosen answer, and its rationale.

Each user then judged, with binary responses, whether (i)~the answer options were valid and meaningful without the question generation context, (ii)~the question was clear and answerable given the question generation context, (iii)~the answering rationale was grounded in the answering context, and (iv)~the chosen answer followed from the rationale.

\Cref{tab:user-study} summarizes the results. The high scores across all criteria support the reliability of using LLMs to generate and answer ADQA questions.
Issues flagged were generally very minor; for instance, 
including ``Verbal'' in the answer options for the question ``Who ultimately executed the task that Keaton hesitated to perform?'' was seen as odd by one user, who may have confused the name of one of the main characters in the movie---``Verbal'', with the common noun (more examples in \cref{tab:userstudy-examples}).

\subsection{ADQA: Context Ablation}
\label{subsec:exp:ablation}
\input{tables/adqa-setup-results}

We present the impact of different context inputs with human narrated AD tracks in \cref{tab:adqa-setup-results}.
The results are reported on all data (private + public).

For Visual Appreciation (VA), we use AudioVault track 1 to create questions, while for answering we use AudioVault track 2 (for CMD-AD) or LSMDC (for MAD-eval).
Narrative Understanding (NU) questions are created using plot descriptions.
For answering NU, we consider both AudioVault tracks for CMD-AD; and AudioVault and LSMDC tracks for MAD-eval.

\paragraph{Prior knowledge.}
LLMs are able to answer a substantial number of questions correctly (CA) with no context (NoCtx), highlighting that \textit{LLMs have extensive prior knowledge and ability to guess based on common sense}.
Interestingly, including the movie name (MN) does not improve results.

\paragraph{Impact of rationale.}
We first examine how generating rationales affects the LLM’s ability to ground its answers in context.
For VA questions on CMD-AD, the baseline accuracy without context (NoCtx) is 55.9\% CA. Adding dialog + AD reduces CA to 41.4\%, and CC further drops to 30.2\%, indicating the LLM struggles when forced to justify answers using context.
For MAD-eval, CC is 59.0\% and hints at a stronger agreement between AV$_1$ and LSMDC. Across both datasets, this confirms that \textit{requiring rationales compels the LLM to ground answers in context} rather than exclusively use prior knowledge.

\paragraph{Context utilization.}
We analyze how different contexts (dialog, AD, or both) are used by the LLM when answering ADQA questions.
Across both datasets and question types, the \textit{share of answers actually grounded in context (AC) rises as richer context is provided}: dialog alone yields low AC for VA (19.6\%, 24.7\%) since ADs provide most visual details; adding ADs raises AC substantially (60.2\%, 75.9\%); dialog+AD improves it further (62.6\%, 76.8\%).
For NU, dialog alone already offers strong narrative cues (81.9\%, 68.1\% AC), while AD-only context is less helpful (73.5\%, 58.7\% AC). Yet, combining dialog+AD further boosts grounding (92.0\%, 86.4\% AC).

\paragraph{Impact of context type.}
Across both datasets and question types, we see a consistent trend: \textit{enriching the answering context improves  accuracy (CC)}.
For VA, dialog-only context leads to lower CC (9.7\% CMD-AD) vs. AD-only (26.5\%), highlighting that ADs convey essential visual details missing from dialog.
For NU, dialog only (61.9\% CC, CMD-AD) outperforms AD-only (56.4\% CC) emphasizing the importance of dialog in driving the story.
However, the best results  consistently come  from dialog + AD with 30.2\% CC for VA and 73.0\% CC for NU on CMD-AD.
MAD-eval shows similar trends, dialog + AD performs best.
These experiments demonstrate that, across both datasets, \textit{ADs significantly enhance visual appreciation} and strongly  \textit{complement dialog for improved narrative understanding}.
They also validate that \textbf{ADQA is well-positioned to evaluate the richness and relevance of generated ADs}.

\subsection{Evaluating AD Generation Methods}
\label{subsec:exp:methods}

\paragraph{Methods.}
We report results for various methods in \cref{tab:adqa-methods-private-movies}.
We evaluate multiple fine-tuned (AutoAD-3~\cite{han2024autoad3}, UniAD~\cite{wang2025uniad}, DistinctAD~\cite{fang2025distinctad}) and zero-shot (AutoAD-0~\cite{xie2024autoad0}, NarrAD~\cite{park2025narrad}, Shot-by-Shot~\cite{xie2025shotbyshot}),  AD generation methods.
We also evaluate dense descriptions obtained from state-of-the-art VideoLMs (Qwen2VL~\cite{wang2024qwen2vl}), prompted to generate paragraph-length rich descriptions relevant for AD generation (disregarding the temporal constraints required for ADs).
Comparison against ground-truth AD sources (AudioVault AV$_1$, AV$_2$, and LSMDC) provide upper bounds for model performance and test the limits of what is realistically achievable on ADQA.
A lower bound is established with context as dialog-only.

\paragraph{Setup.}
For CMD-AD, we evaluate generated ADs against questions obtained for the 93 private movies.
AV$_2$ used in answering is an exception as it only has 17 movies.
For MAD-eval, we evaluate all methods against the 9 private movies.
The CC$_h$ in Accuracy Ratio (\cref{eq:accratio}) is determined as follows.
For VA questions, we use AV$_2$ for CMD-AD and LSMDC for MAD-eval.
For NU questions, we use AV$_1$ for CMD-AD and average accuracy of AV$_1$ and LSMDC for MAD-eval.

\subsubsection{Results}

\subparagraph{Human ADs} outperform all models across both question types and datasets.
On CMD-AD NU, AV$_1$ (72.8\%) and AV$_2$ (75.0\%) achieve similar CC scores.
Likewise, LSMDC (65.2\%) and AV$_1$ (69.5\%) reach similar CC scores on MAD-eval NU, thus validating the benchmark.

\input{tables/adqa-methods-private-movies}

\subparagraph{Dialog-only Baseline.}
For narrative understanding,
dialog-only provides a strong baseline for CMD-AD (CC 59.1\%), which all models surpass except Shot-by-Shot (Qwen2-VL-7B, LLaMA3-8B), and dense descriptions from Q2VL.
In contrast, for MAD-eval, dialog-only achieves CC 50.3\%, and ADs from most models do not surpass this score.
NarrAD that uses scripts stands out with CC 52.4\%, and Shot-by-Shot (GPT-4o, GPT-4o) with CC 51.7\%.
For visual appreciation,
dialog-only is a good baseline for VA (CC 9.8\% on CMD-AD, 11.8\% on MAD-eval) and all models are able to outperform it.

\subparagraph{Movie Scripts.}
NarrAD, which utilizes movie scripts as input, achieves the highest MAD-eval NU score (CC 52.4\%). 
This indicates that movie scripts provide strong cues for generating ADs.
However, as such scripts are not always available, we do not include NarrAD in further comparisons.

\subparagraph{Best Performers.}
Looking at CC,
Shot-by-Shot (GPT-4o, GPT-4o) achieves highest score on CMD-AD (VA 21.1\%, NU 70.2\%) and MAD-eval  (VA 23.5\%, NU 51.7\%).
This indicates that it is indeed important to look beyond the trimmed clips.

Accuracy ratio highlights the wide gap between human top-line performance (second AD track) and model generated ADs.
On VA, MAD-eval (24.8\%) seems harder than CMD-AD (55.4\%), as seen in Shot-by-Shot (GPT-4o, GPT-4o) results.
Furthermore, on MAD-eval for the NU task, generated ADs (except NarrAD and Shot-by-Shot (GPT-4o, GPT-4o)) contribute negatively in assisting narrative understanding.
Interestingly, recent zero-shot methods outperform older trained models.

\subparagraph{Dense Descriptions.}
Q2VL achieves relatively high CMD-AD VA CC (17.1\%). This suggests that VA questions that ask about specific visual details, benefit from richly detailed captions. In contrast, it performs poorly on NU (CC 51.5\%), likely because NU requires broader coherence, and an overload of dense information may hinder comprehension.

\paragraph{Evaluation server.}
We host ADQA as an evaluation server with a leaderboard where participants upload their generated ADs for evaluation on the private set.
This ensures fair comparisons with the same answering prompt and LLM across all methods.
For completeness, evaluation on the public set is reported in \cref{app:benchmark}.

%% file: tables/user-study.tex
\begin{table}[t]
\centering
\small
\tabcolsep=0.18cm
\begin{tabular}{lcc}
\toprule
 & {\color{RoyalPurple}\textbf{Vis App}} & \color{ForestGreen}\textbf{Narr Und} \\
\midrule
Valid options w/o context               & 95.3\% & 95.3\% \\
Question clear and answerable              & 96.9\% & 98.4\% \\
Answer rationale from context       & 93.7\% & 98.4\% \\
Chosen answer from rationale       & 92.2\% & 96.9\% \\
\bottomrule
\end{tabular}
\vspace{-2mm}
\caption{8 participants judged validity of generated questions, and checked whether the chosen answer and its rationale were well grounded.}
\vspace{-2mm}
\label{tab:user-study}
\end{table}

%% file: tables/adqa-setup-results.tex
\begin{table*}[t]
\centering
\small
\tabcolsep=0.14cm
\begin{tabular}{l l ccc cc p{0.5mm} ccc ccc ccc}
\toprule
& \multirow{2}{*}{Dataset} & \multirow{2}{*}{\#M} & \multicolumn{2}{c}{Source} & NoCtx & MN && \multicolumn{3}{c}{Dialog-only} & \multicolumn{3}{c}{AD-only} & \multicolumn{3}{c}{Dialog + AD} \\
\cmidrule(lr){4-5} \cmidrule(lr){9-11} \cmidrule(lr){12-14} \cmidrule(lr){15-17}
& & & Q & A & CA & CA && CA & AC & CC & CA & AC & CC & CA & AC & CC \\
\midrule
{\color{RoyalPurple}Visual}
& CMD-AD    & 17 & AV$_1$ & AV$_2$ & 55.9 & 55.9 && 44.0 & 19.6 & 09.7 & 37.6 & 60.2 & 26.5 & 41.4 & 62.6 & 30.2 \\
{\color{RoyalPurple}Appreciation}
& MAD-eval  & 10 & AV$_1$ & L      & 46.9 & 47.0 && 41.1 & 24.7 & 11.7 & 67.7 & 75.9 & 58.5 & 67.0 & 76.8 & 59.0 \\
\midrule
& CMD-AD    & 17 & plot & AV$_1$  & 60.2 & 59.1 && 70.3 & 81.9 & 61.9 & 70.9 & 73.5 & 56.4 & 76.7 & 92.0 & 73.0 \\
{\color{ForestGreen}Narrative}
& CMD-AD    & 17 & plot & AV$_2$  & -    & -    && 69.2 & 80.8 & 60.5 & 68.9 & 69.6 & 51.8 & 77.8 & 92.0 & 75.0 \\
{\color{ForestGreen}Understanding}
& MAD-eval  & 10 & plot & AV$_1$  & 52.1 & 52.7 && 64.1 & 68.1 & 50.1 & 62.0 & 58.7 & 42.5 & 74.9 & 86.4 & 68.4 \\
& MAD-eval  & 10 & plot & L       & -    & -    && 62.4 & 66.8 & 48.7 & 60.0 & 54.8 & 38.6 & 71.6 & 84.2 & 64.4 \\

\bottomrule
\end{tabular}
\vspace{-2mm}
\caption{%
Ablation results for different types of context.
All sources are human authored ADs.
Source acronyms are
AV$_1$: AudioVault source 1,
AV$_2$: AudioVault source 2,
L: LSMDC, and
plot: narrative description accompanying or aligned to the video segments.
Other acronyms include
\#M: number of movies,
NoCtx: no context, and
MN: context is movie name.
The metrics are
CA: correct answer,
AC: answer uses context,
CC: correct answer using context.
}
\label{tab:adqa-setup-results}
\vspace{-2mm}
\end{table*}

%% file: tables/adqa-methods-private-movies.tex
\begin{table}[t]
\centering
\small
\tabcolsep=0.10cm
\begin{tabular}{@{} ll c cc cccc @{}}
\toprule
& \multirow{2}{*}{Method} & \multirow{2}{*}{\rotatebox{90}{Train}} & \multicolumn{2}{c}{Old Metrics} & \multicolumn{2}{c}{\color{RoyalPurple}\textbf{Vis App}} & \multicolumn{2}{c}{\color{ForestGreen}\textbf{Narr Und}} \\
& & & C & LLMe & CC & Ratio & CC & Ratio \\
\midrule
\multirow{7}{*}{\rotatebox{90}{\textbf{CMD-AD}}} 
& Dialog-only& - & -    & -    & 9.8 & 0 & 59.1 & 0 \\
& AutoAD-III   & \cmark & 25.0 & 2.01 & 14.7 & 24.0 & 63.5 & 32.1 \\
& UniAD*     & \cmark & 21.8 & \textbf{2.92} & 14.1 & 21.0 & 63.2 & 29.9 \\
& AutoAD-Zero   & \xmark & 17.7 & 1.96 & 13.2 & 16.7 & 63.2 & 29.9 \\
& Q2VL       & \xmark & -    & -    & 17.1 & 35.8 & 51.5 & -55.4 \\
& S-by-S-Qwen     & \xmark & \textbf{26.3} & 2.42 & 17.5 & 37.7 & 69.2 & 73.7 \\
& S-by-S-GPT4o     & \xmark & 26.1 & 2.66 & 21.1 & \textbf{55.4} & 70.2 & \textbf{81.0} \\
& CoherentAD    & \xmark & - & - & 17.9 & 39.7 & 67.8 & 63.5 \\
\cmidrule(l){2-9}
& AV$_1$      & - & -    & -    & -    & -    & 72.8 & 100  \\
& AV$_2$ (17) & - & -    & -    & 30.2 & 100  & 75.0 & 116 \\ 
\midrule
\multirow{8}{*}{\rotatebox{90}{\textbf{MAD-eval}}}
& Dialog-only& - & -    & -    & 11.8 & 0 & 50.3 & 0 \\
& AutoAD-III   & \cmark & 24.0 & 2.20 & 14.5 & \phantom{0}5.7 & 43.7 & -38.7 \\
& DistinctAD & \cmark & 27.3 & 2.27 & 13.5 & \phantom{0}3.6 & 37.8 & -73.3 \\
& UniAD*     & \cmark & \textbf{28.2} & 2.46 & 15.8 & \phantom{0}8.5 & 44.0 & -37.0 \\
& AutoAD-Zero   & \xmark & 22.4 & 2.20 & 13.9 & \phantom{0}4.5 & 44.4 & -34.6 \\
& NarrAD     & \xmark & 26.4 & 2.64 & 22.3 & 22.3 & 52.4 & \textbf{12.3} \\
& S-by-S-Qwen     & \xmark & 25.0 & - & 19.6 & 16.6 & 48.9 & -8.2 \\
& S-by-S-GPT4o     & \xmark & 26.9 & - & 23.5 & \textbf{24.8} & 51.7 & \phantom{0}8.2 \\
\cmidrule(l){2-9} 
& AV$_1$     & - & -    & -    & -    & -    & 69.5 & 112  \\
& LSMDC      & - & -    & -    & 58.9 & 100  & 65.2 & 87.4 \\
\bottomrule
\end{tabular}
\vspace{-2mm}
\caption{Evaluation of various generated and human-authored ADs on ADQA private set.
Results are reported with dialog + AD as context.
DistinctAD uses Llama.
NarrAD uses curated ADs.
UniAD has some missing outputs.
The "Train" column indicates if the method is fine-tuned (\cmark) or zero-shot (\xmark).
Acronyms:
{\color{RoyalPurple}Vis App: Visual Appreciation},
{\color{ForestGreen}Narr Und: Narrative Understanding}.
The metrics are
C: CIDEr, LLMe: LLM-AD-eval~\cite{han2024autoad3},
CC: Correct answer using Context, and
Ratio: Accuracy ratio.
}
\vspace{-2mm}
\label{tab:adqa-methods-private-movies}
\end{table}

%% file: sections/7_conclusion.tex
\section{Recommendations for Future Work}
\label{sec:recommend}

While automatic AD generation has made progress, a significant gap remains in supporting visual appreciation and narrative understanding for BLV users.
We outline research directions to bridge this gap:

\textit{1. From clips to videos.}
Generation and evaluation should move to the video level as considering ADs in isolation will not result in coherent descriptions necessary for appreciation and understanding.

\textit{2. Focus on narratives.}
Current models fail to connect individual events into a story, often producing repetitive ADs.
Combining ADs with dialog and training models to answer ADQA-like questions may help.

\textit{3. VLMs hold promise.}
Dense Q2VL results suggest VLMs can extract rich details.
Future work may continue to focus on distilling these into concise ADs that fit naturally within dialog.

\textit{4. Scripts, with caution.}
NarrAD results show that scripts provide useful cues, but they aren’t always available or aligned with the final cut.
Relying on scripts may limit real-world adoption.

\textit{5. More hands-on evaluation.}
Overreliance on LLMs risks leakage of prior knowledge or ungrounded rationales.
A holistic evaluation that follows the AD guidelines is needed.

\section{Conclusion}
\label{sec:conclusion}

We proposed ADQA, a new evaluation paradigm for automatic AD generation methods that addresses   two themes central to ADs, whether they help BLV users with \textit{visual appreciation} and \textit{narrative understanding}.
To analyze the subjectivity of ADs, we aligned and compared two human-narrated AD tracks, revealing issues with current video captioning-like setups, semantic similarity based metrics like BERT similarity, and n-gram based metrics like CIDEr.
The second track also provided a top-line human performance on ADQA.
Evaluation of current AD generation methods showed a large gap to human-authored ADs.
We also provided several recommendations for future work based on our findings.

\paragraph{Acknowledgments.}
This project was supported by funding from SERB SRG/2023/002544 and an Adobe Research India gift.
We thank Google for sponsoring GCP credits that supported the use of Gemini API.
We thank volunteers for participating in the user study and all AD researchers that provided their model outputs.

%% file: sections/8_limitations.tex
\section{Limitations}
\label{sec:limitations}

While ADQA introduces a narrative-aligned framework for evaluating ADs, some limitations are:

\textit{1. Temporal feasibility not evaluated.}
We do not currently assess whether model-generated ADs are too long to fit naturally between dialog segments.
A naive solution would be to cap the AD length based on available dialog-free intervals (\eg~assuming 160 words-per-minute~\cite{visual_made_verbal}).
However, this may encourage models to saturate silent gaps with verbose descriptions, potentially overwhelming the viewer and detracting from the cinematic experience.

\textit{2. Variability in LLM outputs.}
Language model outputs are inherently non-deterministic, introducing small fluctuations in evaluation scores across runs. 
While this randomness is small, it raises some reproducibility concerns.
We plan to manage this by running evaluations through a server where participants can upload their generated AD predictions with a rate limit.

\textit{3. Prior knowledge leakage.}
Despite best efforts to mitigate answering via prior knowledge, LLMs possess extensive information about popular movies.
However, as the same model is used to assess all automatic AD generation methods and human-authored ADs, the relative scores remain valid, even if absolute performance may be inflated.

%% file: appendix/alignment.tex
\section{Responsible NLP Details}

\paragraph{Model Size and Computational Budget:}
For computing sentence level similarity, we use \textit{bert-base-uncased}, a 110M parameter model trained on English language.
The bulk of work was done using Gemini APIs, with an approximate cost of generating questions $< \$100$ and evaluation of each AD generation method at $< \$5$ per experiment.

\paragraph{License or Terms of use for Artifacts:}
Movie video clips from CMD are available on YouTube for public access. 
AudioVault is a non-commercial entity that hosts audio-only movie files for the benefit of Blind and Low Vision individuals.
We use AudioVault to access audio descriptions.

We made use of AI assistants such as ChatGPT~\cite{chatgpt} to help with coding. 
Commercial AI models Gemini were used for experiments and data creation.
We use Gemini-2.5-Pro~\cite{gemini25pro} for generating all questions in the dataset, and use Gemini-2.0-Flash~\cite{gemini2flash} for answering the questions.
Gemini API Terms of Service allows the API for commercial or research use.

\paragraph{Miscellaneous:}
Movie data may contain offensive or explicit content.
All experimental results are presented for a single run.

\section{Details of the AD Mapping Process}
\label{app:alignment}

Post-alignment of AD + dialog transcriptions, we map ADs from $T_1$ (track 1) and $T_2$ (track 2) using the process defined below:

\begin{enumerate}[nosep]
\item Consider an AD $A^1_i$ in $T_1$ with duration $d^1_i$.
\item Identify the slope relevant to this AD (for some movies the offset changes across the movie due to censored scenes).
\item Predict the time duration $\hat{d}^2_i$ in $T_2$ for this AD using the slope and offset, and add a \SI{1}{\second} buffer on each side.
\item Get a list of ADs in $T_2$ having overlap with $\hat{d}^2_i$.
For each AD, $A^2_j$ with duration $d^2_j$:
\begin{enumerate}[nosep]
\item Compute overlap score $O(A^1_i, A^2_j)$ as:
\begin{equation}
O(A^1_i, A^2_j) = \frac{\cap(\hat{d}^2_i, d^2_j)}{\min(\hat{d}^2_i, d^2_j)} \, .
\end{equation}
\item If $O > 50\%$, create a mapping. 
\end{enumerate}
\item Repeat above process (steps 1-5) for all ADs $A^1_i$ in $T_1$.
\item Also repeat above process (steps 1-5) for all ADs $A^2_j$ in $T_2$.
\begin{enumerate}[nosep]
\item If $A^2_j$ in $T_2$ fails to find a match and was not already mapped in steps 1-4, then we consider $A^2_j$ as non-aligned.
\end{enumerate}
\item All ADs $A^1_i$ in $T_1$ without a mapping to any AD in $T_2$ are also considered non-aligned.
\end{enumerate}

%% file: appendix/benchmark.tex
\section{Benchmark}
\label{app:benchmark}

\subsection{Public and Private Subsets}
To facilitate understanding of the benchmark, we publicly release all questions and answers from 5 movies in CMD and 1 movie in MAD-eval. These movies were selected to be among the most representative of the full benchmark, capturing both model ranking and absolute performance. The rest of the 93 CMD movies and 9 MAD-eval movies form the private testing set to be used for evaluation in the leaderboard. \cref{tab:adqa-public-private-comparison} shows model outputs on the entire dataset and public and private sets.

\input{tables/adqa-public-private-comparison}

\subsection{Leaderboard}
An online leaderboard can be accessed through the project website \url{https://katha-ai.github.io/projects/adqa/}.
Researchers can submit their model generated ADs for evaluation on the private testing set with rate limits to prevent overfitting.

%% file: tables/adqa-public-private-comparison.tex
\begin{table}[t]
\centering
\small
\setlength{\tabcolsep}{4pt}
\begin{tabular}{@{}llcc|cc|cc@{}}
\toprule
& \multirow{2}{*}{Model} 
& \multicolumn{2}{c|}{\textbf{All}} 
& \multicolumn{2}{c|}{\textbf{Private}} 
& \multicolumn{2}{c}{\textbf{Public}} \\
& & {\color{RoyalPurple}\textbf{VA}} & {\color{ForestGreen}\textbf{NU}} 
  & {\color{RoyalPurple}\textbf{VA}} & {\color{ForestGreen}\textbf{NU}} 
  & {\color{RoyalPurple}\textbf{VA}} & {\color{ForestGreen}\textbf{NU}} \\
\midrule
\multirow{6}{*}{\rotatebox{90}{CMD-AD}}
& Dialogue-only & 10.0 & 58.9 & 9.8 & 59.1 & 13.8 & 54.9 \\
& AutoAD-0      & 13.4 & 62.9 & 13.2 & 63.2 & 15.5 & 56.9 \\
& AutoAD-III    & 14.9 & 63.2 & 14.7 & 63.5 & 18.0 & 56.9 \\
& Qwen2         & 17.2 & 51.2 & 17.1 & 51.5 & 17.9 & 45.8 \\
& UniAD*        & 14.3 & 63.0 & 14.1 & 63.2 & 17.8 & 60.1 \\
& S-by-S-Qwen   & 17.7 & 69.1 & 17.5 & 69.2 & 21.2 & 65.4 \\
& S-by-S-GPT4o  & 21.1 & 70.1 & 21.1 & 70.2 & 20.3 & 67.3 \\
& CoherentAD    & 18.0 & 67.7 & 17.9 & 67.8 & 18.9 & 66.0 \\
\cmidrule(l){2-8}
& AV$_1$ (GT)   & --   & 72.7 & --   & 72.8 & --   & 71.9 \\
& AV$_2$ (17)   & 30.2 & 75.0 & 30.2 & 75.0 & --   & --   \\
\midrule
\multirow{7}{*}{\rotatebox{90}{MAD-eval}}
& Dialogue-only & 11.7 & 48.7 & 11.8 & 50.3 & 11.0 & 33.7 \\
& AutoAD-0      & 14.3 & 44.2 & 13.9 & 44.4 & 19.1 & 42.1 \\
& AutoAD-III    & 14.7 & 42.9 & 14.5 & 43.7 & 17.5 & 34.7 \\
& DistinctAD    & 13.7 & 37.6 & 13.5 & 37.8 & 16.1 & 35.3 \\
& NarrAD        & 22.7 & 52.7 & 22.3 & 52.4 & 26.7 & 54.7 \\
& UniAD*        & 16.0 & 43.4 & 15.8 & 44.0 & 18.5 & 38.4 \\
& NarrAD        & 22.7 & 52.7 & 22.3 & 52.4 & 26.7 & 54.7 \\
& S-by-S-Qwen   & 19.8 & 48.4 & 19.6 & 48.9 & 23.2 & 44.2 \\
& S-by-S-GPT4o  & 24.0 & 51.5 & 23.5 & 51.7 & 29.8 & 48.9 \\
\cmidrule(l){2-8}
& AV$_1$        & --   & 68.4 & --   & 69.5 & --   & 57.9 \\
& LSMDC         & 59.0 & 64.4 & 58.9 & 65.2 & 60.0 & 57.4 \\
\bottomrule
\end{tabular}
\caption{Model performance on subsets of the benchmark. 
Columns report results on the full set (\textbf{All}), the hidden leaderboard set (\textbf{Private}), 
and the publicly released set (\textbf{Public}).  
{\color{RoyalPurple}\textbf{VA}} = Visual Appreciation, 
{\color{ForestGreen}\textbf{NU}} = Narrative Understanding.  
For CMD-AD, $n=\{98,93,5\}$ correspond to All/Private/Public respectively;  
for MAD-eval, $n=\{10,9,1\}$. 
Note that AV$_2$ has no movies in the public set and hence no public results are reported.
Scores are Correct answer using Context (CC) ($\uparrow$).}
\label{tab:adqa-public-private-comparison}
\end{table}

%% file: appendix/qualitative.tex
\section{Qualitative}
\label{app:qualitative}

\input{tables/supp-adqa_questions}

\paragraph{Examples from ADQA.}
\label{app:qualitative:adqa}
\cref{tab:supp-adqa_questions} presents several questions from the ADQA benchmark along with multiple choice options, the correct answer, and the rationale for the answer.
We observe that the rationale often refers to specific parts of the AD (for appreciation QAs) or the plot-like descriptions (for understanding QAs).

\paragraph{Generated ADs and Answering ADQA.}
\label{app:qualitative:answering}
\cref{tab:supp-adqa_answering} presents some examples of LLM based answering for different input contexts.
We see that dialog + AD results in meaningful rationales indicating that the model is able to evaluate the quality of AD provided as context.

\input{tables/supp-adqa_answering}

%% file: tables/supp-adqa_questions.tex
\begin{table*}[t]
\centering
\small
\tabcolsep=0.2cm
\begin{tabular}{l p{4.1cm} p{5.5cm} p{4.5cm}}
\toprule
\# &
\multicolumn{1}{c}{Question} &
\multicolumn{1}{c}{Answer Options} &
\multicolumn{1}{c}{Rationale} \\
\midrule
\multicolumn{4}{c}{\color{RoyalPurple}\textbf{Visual Appreciation}} \\
\midrule
1 & How many companions accompany Biff? &
A) One\newline
B) Two\newline
\textbf{C) Three}\newline
D) Four\newline
E) Five &
As specified in the audio description, Biff has three buddies with him. \\
\midrule
2 & What physical characteristics are noted about Biff? &
A) Short and slim with long hair\newline
B) Average height and build with curly hair\newline
\textbf{C) Tall and muscular with short hair}\newline
D) Short and stocky with blond hair\newline
E) Tall and thin with slicked-back hair &
As specified in the audio description, Biff is tall and muscular and wears his hair cut short. \\
\midrule
3 & What is the location of the vehicle Biff and his buddies enter? &
A) In the cafe's parking lot\newline
B) Down the street\newline
C) In an alleyway\newline
\textbf{D) Parked outside the cafe}\newline
E) Across the town square &
As specified in the audio description, the black convertible is parked outside the cafe. \\
\midrule
4 & What type of eyewear is another of Biff's buddies wearing? &
A) Sunglasses\newline
B) Reading glasses\newline
C) Safety goggles\newline
\textbf{D) 3D glasses}\newline
E) A monocle &
As specified in the audio description, another buddy has 3D glasses. \\
\midrule
5 & What simultaneous action do Marty and the nearby kid perform? &
A) They both sigh\newline
\textbf{B) They both turn their heads}\newline
C) They both check their watches\newline
D) They both take a drink\newline
E) They both look down &
As specified in the audio description, both boys turn their heads. \\
\midrule
\multicolumn{4}{c}{\color{ForestGreen}\textbf{Narrative Understanding}} \\
\midrule
6 & What distressing situation involving his young father does Marty witness at this location after they meet? &
A) His father failing an important test\newline
B) His father being rejected for a date\newline
\textbf{C) His father being subjected to bullying}\newline
D) His father getting into a car accident\newline
E) His father being scolded by his parents &
The description indicates Marty 'watches him get bullied by Biff.' \\
\midrule
7 & Who is primarily responsible for the mistreatment directed towards Marty's young father during the scene at the diner? &
A) Strickland, the school principal\newline
B) Marty, through an accidental intervention\newline
\textbf{C) Biff Tannen}\newline
D) A group of his father's classmates\newline
E) The diner's cook &
The description explicitly states that Marty's father is bullied 'by Biff.' Biff Tannen is the character's full name. \\
\midrule
8 & Whom does Marty specifically meet at the diner before witnessing an incident involving Biff? &
A) His young mother, Lorraine\newline
\textbf{B) His young father, George}\newline
C) A younger version of Doc Brown\newline
D) His older brother, Dave, as a teen\newline
E) Goldie Wilson, the future mayor &
The description states, 'Marty meets his young father at the diner...'. George is the name of Marty's father. \\
\midrule
9 & What is Marty's primary role concerning the confrontation involving his young father and Biff at the diner? &
A) He directly intervenes to help his father.\newline
B) He actively encourages Biff.\newline
\textbf{C) He observes the situation unfold.}\newline
D) He calls for help from others in the diner.\newline
E) He tries to create a distraction. &
The description states, 'Marty ... watches him get bullied by Biff,' indicating his role as an observer of the event. \\
\midrule
10 & To what year is Marty transported after accelerating the DeLorean under duress? &
A) 1885\newline
\textbf{B) 1955}\newline
C) 1985\newline
D) 2015\newline
E) An unknown future year. &
The description specifies that Marty "goes back in time to 1955." \\
\bottomrule
\end{tabular}
\caption{Example multiple-choice question-answers for {\color{RoyalPurple}\textbf{Visual Appreciation}} (top) and
{\color{ForestGreen}\textbf{Narrative Understanding}} (bottom) from ADQA.
The correct answer is highlighted in bold.}
\label{tab:supp-adqa_questions}
\end{table*}

%% file: tables/supp-adqa_answering.tex
\begin{table*}[t]
\centering
\small
\tabcolsep=0.2cm
\begin{tabular}{p{1.8cm} p{5cm} p{6.5cm} c}
\toprule
\multicolumn{1}{c}{Context Type} &
\multicolumn{1}{c}{Answer Options} &
\multicolumn{1}{c}{Rationale} &
\multicolumn{1}{c}{AC} \\
\midrule
\multicolumn{4}{l}{\textbf{Question 1:} How many companions accompany Biff?} \\
\midrule
No Context &
A) One\newline
B) Two\newline
\textcolor{ForestGreen}{\textbf{C) Three}}\newline
D) Four\newline
E) Five &
Bullies are often accompanied by multiple companions for intimidation purposes
 & -\\
 
\midrule
Dialog &
A) One\newline
\textcolor{Red}{\textbf{C) Two}}\newline
\textcolor{ForestGreen}{\textit{C) Three}}\newline
D) Four\newline
E) Five &
The dialogue references Biff and 'those boys', suggesting more than one companion, however without additional information on the exact number of companions accompanying Biff.
 & \xmark\\
 
\midrule
Dialog + AD &
A) One\newline
B) Two\newline
\textbf{\textcolor{ForestGreen}{C) Three}}\newline
D) Four\newline
E) Five &
The audio description says, "The young Beth and three cronies leave, jumping into an open-top car parked outside."
 & \cmark\\

\midrule
\multicolumn{4}{p{13.7cm}}{\textbf{Question 7:} Who is primarily responsible for the mistreatment directed towards Marty's young father during the scene at the diner?} \\
\midrule
No Context &
A) Strickland, the school principal\newline
B) Marty, through an accidental intervention\newline
\textcolor{ForestGreen}{\textbf{C) Biff Tannen}}\newline
D) A group of his father's classmates\newline
E) The diner's cook &
In Back to the Future, Biff Tannen and his gang are the primary antagonists and bullies. It is Biff who instigates and leads the mistreatment of Marty's father, George McFly, at the diner.
 & -\\

\midrule
Dialog &
A) Strickland, the school principal\newline
B) Marty, through an accidental intervention\newline
\textcolor{ForestGreen}{\textbf{C) Biff Tannen}}\newline
D) A group of his father's classmates\newline
E) The diner's cook &
The dialogues "What are you looking at, butthead?", "Hey, Biff, get a hold of this guy's life preserver.", "Dork thinks he's gonna drown.", "So, uh, how about my homework, McFly?", "Oh, McFly, your shoe's untied.", "Don't be so gullible, McFly.", "I don't want to see you in here again." all indicate that Biff Tannen is mistreating George McFly.
 & \cmark\\

\midrule
Dialog + AD &
A) Strickland, the school principal\newline
B) Marty, through an accidental intervention\newline
\textcolor{ForestGreen}{\textbf{C) Biff Tannen}}\newline
D) A group of his father's classmates\newline
E) The diner's cook &
The audio descriptions and dialogue clearly indicate that Biff Tannen is the primary instigator of the mistreatment towards George McFly. Biff is shown interacting with George in a demeaning way, demanding his homework, tripping him, and generally bullying him, as stated by "Dialogue: Well, now, of course not, Biff.", "Dialogue: So, uh, how about my homework, McFly?", "Dialogue: Oh, McFly, your shoe's untied." and "Dialogue: Don't be so gullible, McFly."
 & \cmark\\
\bottomrule
\end{tabular}
\caption{Answering questions 1 and 7 (from \cref{tab:supp-adqa_questions}) using the second AD source for example question from visual appreciation (top) and narrative understanding (bottom).
We show the question and answering using different context types (col 1).
Ground-truth correct answer is highlighted in green while the model predicted answer is in bold (red when wrong, green when correct).
We see that the LLM is able to answer the question even without any context based on common sense reasoning.
However, our process to check whether the context was used to create the rationale helps us assess whether the answer was generated using context (last column, AC).
}
\label{tab:supp-adqa_answering}
\end{table*}

%% file: appendix/userstudy.tex
\section{User Study}
\label{app:userstudy}
We ask users to rate generated questions and answering capabilities of the LLM on 4 criteria.
The results are shared in \cref{tab:user-study}.

\paragraph{Options valid without question generation context.}
You are given the generated question and answer options.
Mark 1 if the options make sense, given the question. Otherwise mark 0.
As long as the options make sense and are not wildly out of distribution you can mark 1.

\begin{itemize}[noitemsep, topsep=0pt]
\item Example where 0 should be marked: 
\begin{itemize}[noitemsep, topsep=0pt]
\item Question: How many coconuts were there on the tree?
\item Options A) 3 B) 2 C) 238746 (too large)
\end{itemize}
\item Example where 1 should be marked:
\begin{itemize}[noitemsep, topsep=0pt]
\item Question: How many coconuts were there on the tree?
\item Options A) 3 B) 2 C) 7 D) 10
\end{itemize}
\end{itemize}

\paragraph{Question clear and answerable from question generation context.}
You are given the question context, generated question, answer options and correct answer.
Mark 1 if question is clear and answerable from the question generation context, otherwise 0.

\begin{itemize}[noitemsep, topsep=0pt]
\item Context:
\begin{itemize}[noitemsep, topsep=0pt]
\item Dialogue: The bear is going to attack us!
\item AD: The bear saunters away.
\end{itemize}
\item Example where 0 should be marked:
\begin{itemize}[noitemsep, topsep=0pt]
\item Question: What color was the bear?
\item Options: A) Brown B) Black C) White
\item Reason: The correct answer is not derivable, given the context.
\end{itemize}
\item Example where 1 should be marked: 
\begin{itemize}[noitemsep, topsep=0pt]
\item Question: What action did the bear take?
\item Options: A) The bear attacked B) The bear stood on its hind legs C) The bear sauntered away.
\item Reason: The correct answer is derivable, given the context.
\end{itemize}
\end{itemize}

\paragraph{Answer rationale is derived from answering context.}
You are given the QA, answer rationale, and the answering context.
Mark 1 if the answer rationale is correctly derived from the answering context, otherwise 0.

\begin{itemize}[noitemsep, topsep=0pt]
\item Answering context: 
\begin{itemize}[noitemsep, topsep=0pt]
\item Audio Description: Mark runs through a series of long hallways.
\item Audio Description: He stops before a large door.
\end{itemize}
\item Example where 1 should be marked:
\begin{itemize}[noitemsep, topsep=0pt]
\item Answer rationale: The audio description clearly states that Mark stops before a large door
\end{itemize}
\item Example where 0 should be marked:
\begin{itemize}[noitemsep, topsep=0pt]
\item Answer rationale: Although the color of the hallway is not explicitly stated, In Severence, hallways are famously white.
\end{itemize}
\end{itemize}

\paragraph{Answer is derived from Answer rationale.}
You are given QA, the model prediction, and the answer rationale.
Do not worry about whether the answer rationale is based on the context.
Mark 1 if the correct/generated answer is derived from the answer rationale, otherwise 0.

\begin{itemize}[noitemsep, topsep=0pt]
\item Example where 1 should be marked:
\begin{itemize}[noitemsep, topsep=0pt]
\item Answer rationale: Although the color of the hallway is not explicitly stated, In Severence, hallways are famously white.
\item Correct/generated answer: White
\end{itemize}
\item Example where 0 should be marked:
\begin{itemize}[noitemsep, topsep=0pt]
\item Answer rationale: Although the height of Burj Khalifa is not stated, skyscrapers are generally 100 meters.
\item Correct/generated answer: 830m
\end{itemize}
\end{itemize}

\input{tables/userstudy-examples}

%% file: tables/userstudy-examples.tex
\begin{table*}[t]
\centering
\small
\tabcolsep=0.15cm
\begin{tabular}{p{4.5cm} p{4.2cm} p{1.3cm} p{4cm}}
\toprule
\textbf{Question} & 
\textbf{Options} & 
\textbf{Response} &
\textbf{Potential reason for marking False} \\
\midrule

In this situation, who ultimately executed the task that Keaton hesitated to perform? &
A) The individual eventually complied\newline
B) Another, unnamed character\newline
\textbf{\textcolor{ForestGreen}{C) Verbal}}\newline
D) Keaton, after a delay\newline
E) The opportunity was lost &
False &
Likely confused by “Verbal” as a name. \\
\midrule

Whose marriage to Greg is contingent upon the outcome of Jack's lie detector test? &
A) Greg's sister\newline
B) A distant relative of Jack\newline
\textbf{\textcolor{ForestGreen}{C) Jack's daughter}}\newline
D) A mutual friend\newline
E) Jack's ex-wife &
False &
Participant may have assumed the option "Greg's sister" as unlikely. \\
\midrule

How does Edward first observe the scene after the shooting? &
A) By opening a door slightly\newline
B) By looking through a window\newline
\textbf{\textcolor{ForestGreen}{C) By peeping around the corner}}\newline
D) By using a periscope\newline
E) By stepping out fully &
False &
Unknown. \\
\midrule

What type of vehicle does Nikki observe arriving? &
A) Bourne's motorcycle\newline
B) A police car\newline
C) An ambulance\newline
\textbf{\textcolor{ForestGreen}{D) A van}}\newline
E) The asset's hatchback &
False &
"Asset's hatchback" may have seemed strange without context about "The Asset" — a character in the movie.  \\
\midrule

What action does the pursuing individual take towards Bourne? &
A) Shouts warnings\newline
B) Fires a weapon\newline
C) Rides a motorcycle\newline
\textbf{\textcolor{ForestGreen}{D) Runs}}\newline
E) Throws an object &
False &
Unknown. \\
\midrule

What object does Bourne discharge towards the van? &
A) A bullet\newline
B) A flare\newline
\textbf{\textcolor{ForestGreen}{C) A gas canister}}\newline
D) A grappling hook\newline
E) Water from the hose &
False &
Likely found the verb “discharge” odd for all options, one normally shoots a bullet, throws a canister, or fires a flare.\\
\bottomrule
\end{tabular}
\caption{Subset of user study questions where participants judged whether the options were valid given only the question and multiple-choice options (without question generation context).
The correct answer is highlighted in \textbf{\textcolor{ForestGreen}{green and bold}}.
The final column lists potential reasons for incorrect judgments (hypothetical, since we did not collect explicit justifications).} 
\label{tab:userstudy-examples}
\end{table*}

%% file: appendix/prompts.tex
\section{Prompts}
\label{app:prompts}

We provide multiple prompts used throughout this work.
When not mentioned otherwise, we use Gemini 2.5 Pro (\texttt{gemini-2.5-pro-preview-03-25}) for this task.

\begin{enumerate}[noitemsep, topsep=0pt]
\item \cref{fig:ad_vs_dialog} presents the prompt used to classify a WhisperX~\cite{bain2023whisperx} transcription into AD or dialog.
\item \cref{fig:mad_video_splitting} presents the prompt used to align the plot synopses sentences with a dialog + AD movie "script" (not the real script).
\item \cref{fig:appreciation_q} presents the prompt used to generate \textit{visual appreciation} MCQAs in our ADQA benchmark.
\item \cref{fig:understanding_q-cmdad} presents the prompt used to generate \textit{narrative understanding} MCQAs for the CMD-AD dataset.
\item \cref{fig:understanding_q_mad} presents the prompt used to generate \textit{narrative understanding} MCQAs for the MAD-eval dataset.
\item \cref{fig:answering} presents the prompt used to answer questions from ADQA based on dialog and/or AD inputs.
All answering is performed using Gemini-2.0-Flash.
\end{enumerate}

\input{appendix/prompts/ad-vs-dialog}

\input{appendix/prompts/mad-video-splitting}

\input{appendix/prompts/appreciation_q}

\input{appendix/prompts/understanding_q-cmdad}

\input{appendix/prompts/understanding_q-mad}

\input{appendix/prompts/answering}

%% file: appendix/prompts/ad-vs-dialog.tex
\begin{figure*}[t]
\centering
\small
\noindent\begin{minipage}{1.0\linewidth}
\mdfsetup{%
middlelinewidth=1pt,
backgroundcolor=green!3,
innerleftmargin=0.5cm,
innerrightmargin=0.5cm,
roundcorner=15pt}
\begin{mdframed}
\vspace{0.2em}
You are an expert in analyzing movie scripts. You will be given a list of sentences that appear sequentially in a movie. For each sentence, classify it as either:

\begin{itemize}[noitemsep,topsep=0pt]
    \item "dialogue" — if it's a spoken line by a character. Music, background chatter, or anything that is not an audio description should also be classified as dialogue
    \item "AD" — if it's an audio description of what is happening on screen.
\end{itemize}
\vspace{0.2cm}

\textbf{Further description}: \\
\textbf{Dialogue}: Spoken lines from a movie, typically involving characters talking.\\
    Example characteristics:
        \begin{itemize}[noitemsep,topsep=0pt]
            \item Use of first-person pronouns like "I," "me," or "my."
            \item Often conversational or emotional in tone. Could be a command, exclamation, rambling, etc.
            \item Examples:

            \begin{itemize}[noitemsep,topsep=0pt]
                \item "You fought a bear? Are you insane?"
                \item "It was either me or him. And honestly, I think I was more scared than he was."
                \item "What even possessed you to go into the forest alone?"
                \item "Get up"
                \item "Move"
            \end{itemize}
        \end{itemize}

\vspace{0.2cm}

\textbf{Audio Description (AD)}: Sentences that narrate the visual elements of a movie, intended for blind or visually impaired viewers.\\
    Example characteristics:
        \begin{itemize}[noitemsep,topsep=0pt]
            \item Is a narrator describing the scene visually.
            \item Descriptive and neutral tone.
            \item Often focuses on actions, settings, or appearances.
            \item Usually starts describing a scene by setting up the environment like "Outside", "Downstairs", "In a sunny afternoon outside", "Out in the snow", etc.
            \item No first-person perspective or conversational cues.
            \item Examples:
            \begin{itemize}[noitemsep,topsep=0pt]
                 \item "A dense Russian forest, snow falls steadily, blanketing the ground. A man steps forward, his breath visible in the icy air."
                \item "The man lunges at the bear with a crude spear, but the bear swats it aside effortlessly."
                \item "He gets up"
            \end{itemize}
        \end{itemize}

\vspace{0.2cm}

\textbf{Instructions:}
\begin{enumerate}[noitemsep,topsep=0pt]
    \item For \textit{every} input sentence, return exactly one classification: either ``\{dialogue\_tag\}'' for dialogue or ``\{ad\_tag\}'' for Audio Description.
    \item Do \textit{not} skip any inputs, even if they are very short or ambiguous.
    \item Match the output count to the input count. If $n$ sentences are given, return exactly $n$ outputs.
    \item Do \textit{not} add any commentary, explanation, or extra lines. Just one output per sentence: ``\{dialogue\_tag\}'' or ``\{ad\_tag\}''.
    \item Use context between sentences if helpful, since these sentences are sequential from a movie.
    \item Some of the movies may be rated for adult audiences and might contain explicit sentences. This makes no difference; the classification should be done regardless just as for any other sentence. Be careful not to include any unsafe or overly sexual content.
    \item Some sentences might be a mix of AD and dialogue due to transcription errors. These sentences should be labelled ``\{ad\_tag\}'' if the audio description part is more prominent in the sentence, otherwise ``\{dialogue\_tag\}''.
\end{enumerate}

\vspace{0.2cm}

\textbf{Input format}:
\begin{enumerate}[noitemsep, topsep=0pt]
\item \{sentence1\}
\item \{sentence2\}
\item \{sentence3\}
\item ...
\end{enumerate}

\vspace{0.2cm}

\textbf{Output format}:
\begin{enumerate}[noitemsep, topsep=0pt]
    \item \{classification1\}
    \item \{classification2\}
    \item \{classification3\}
    \item ...
\end{enumerate}

\vspace{0.2cm}

\textbf{Here is the input}: \{input\}

\end{mdframed}
\end{minipage}
\caption{Prompt to classify transcriptions into "dialogue" or "AD".}
\label{fig:ad_vs_dialog}
\end{figure*}

%% file: appendix/prompts/mad-video-splitting.tex
\begin{figure*}[t]
\centering
\small
\noindent\begin{minipage}{1.0\linewidth}
\mdfsetup{%
middlelinewidth=1pt,
backgroundcolor=green!3,
innerleftmargin=0.5cm,
innerrightmargin=0.5cm,
roundcorner=15pt}
\begin{mdframed}
\vspace{0.2em}

You are a movie editing AI assistant who's job is to segment the movie script into distinct scenes. Each scene is a self contained logical segment of the movie.
You will be provided with two inputs: Movie script and Plot synopsis. \\

\textbf{Movie script format}:
{\fontsize{8}{9}\selectfont
\begin{verbatim}
Line <number>
<start time in hh:mm:ss.ss> --> <end time in hh:mm:ss.ss>
<Sentence type Dialogue or Audio Description>: <sentence>
\end{verbatim}
}
\textbf{Plot synopsis format}:
<plot synopsis paragraph>

\textbf{Instructions}:
\begin{itemize}[noitemsep,topsep=0pt]
    \item Segment the script into logical scenes, each spanning approximately few minutes of screen time (based on timestamps or logical transitions in dialogue and descriptions).
    \item For each scene, list the index range of script lines (e.g., 1–10, 11–18, etc.).
    \item For each scene, identify which sentence(s) from the plot synopsis match the scene's events, if any. If a scene doesn't match any part of the synopsis, note that no match was found.
    \item Use timestamps, audio descriptions, and dialogue shifts to define scene boundaries.
    \item If two script segments are logically distinct (e.g., a sudden change in location or topic), treat them as separate scenes.
    \item Pay special attention to changes in scenes described in audio descriptions.
    \item If a plot synopsis line spans across multiple consecutive scenes, then merge the scenes into one.
    \item Every detail in the plot synopsis should be explainable from the scene. If some detail exists in another consecutive scene, then merge the scenes.
    \item A scene may have one, multiple, or no corresponding synopsis lines.
    \item Every plot line must be associated to some scene, and each line can only be associated to at most 1 scene.
\end{itemize}

\textbf{Output format}:
{\fontsize{7}{8}\selectfont
\begin{verbatim}
[
    (<Line number of scene start>, <Line number of scene end>, <Plot synopsis sentence(s) that correspond to the scene OR None>),
    (<Line number of scene start>, <Line number of scene end>, <Plot synopsis sentence(s) that correspond to the scene OR None>),
    ...
]
\end{verbatim}
}

\textbf{Input}:\\
\textbf{Movie Script}:
\{movie\_script\}

\textbf{Plot synopsis}:
\{plot\_synopsis\}

\textbf{Output}:

\end{mdframed}
\end{minipage}

\caption{Prompt used to align the plot synopses sentences with a dialog + AD movie "script" (not the real script).}
\label{fig:mad_video_splitting}
\end{figure*}

%% file: appendix/prompts/appreciation_q.tex
\begin{figure*}[t]
\centering
\small
\noindent\begin{minipage}{1.0\linewidth}
\mdfsetup{%
middlelinewidth=1pt,
backgroundcolor=green!3,
innerleftmargin=0.5cm,
innerrightmargin=0.5cm,
roundcorner=15pt}
\begin{mdframed}
\vspace{0.2em}

You are given a movie scene in text form, which consists of dialogues and audio descriptions. Your task is to generate questions exclusively based on the audio descriptions, ignoring the dialogues and only using them for context.\\
Every audio description sentence has to be used to construct 1 or more questions asking about direct facts. 
The questions must ask about every factual detail about the audio description sentence.\\

\textbf{Examples}:
    An AD sentence such as "A green truck speeds through the highway crossing a yellow barrier" can become multiple questions such as "What vehicle is seen on the highway?", "What is the color of the vehicle going on the highway?", "What can be said about the speed of the vehicle on the highway?", "What does the vehicle cross on the highway?", "What color is the barrier on the highway?", etc.
    Audio descriptions that are used to establish the scene such as "Outside", "Later that night", "Inside the home", "Now, inside" can be converted to questions about the scene "Where is the scene taking place?", "What time of day is the scene taking place?", etc. \\

\textbf{Question and Answer Format}:\\
\textbf{Questions}: Limited to one or two lines, formulated to be insightful and not overtly indicative of the answer. Avoid using overly descriptive language that could hint at the correct answer. If there are no good questions to be generated, return an empty json.\\
\textbf{Answers}: Five options per question, formatted as "- A), - B), - C), - D), and - E)", concise and reflective of the question’s depth.\\
\textbf{Answer Key}: Specify the correct answer clearly with the formatting, "Correct Answer:", in the line following all the answer options.\\
\textbf{Rationale}: Write a rationale explaining the correctness of the "Answer Key" based on the scene’s context in the next line.\\
The response should be in json format without any extra comments.\\

\textbf{Very Important Rule}: Make sure none of the question is answerable by looking at other questions and their options. \\

\textbf{Output format}:\\
Return json formatted text. Example:\\
{\fontsize{7}{8}\selectfont
\begin{verbatim}
[
    {
        "question": "question text1",
        "options": ["A) answer key 1", "B) answer key 2", "C) answer key 3", "D) answer key 4", "E) answer key 5"],
        "correct_answer": "E) answer key 5",
        "rationale": "As specified in the audio description, <rationale>",
    },
    {
        "question": "question text1",
        "options": ["A) answer key 1", "B) answer key 2", "C) answer key 3", "D) answer key 4", "E) answer key 5"],
        "correct_answer": "A) answer key 1",
        "rationale": "As specified in the audio description, <rationale>",
    }
]
\end{verbatim}
}

\end{mdframed}
\end{minipage}
\caption{Prompt used to generate \textcolor{RoyalPurple}{\textbf{visual appreciation}} MCQAs in our ADQA benchmark.}
\label{fig:appreciation_q}
\end{figure*}

%% file: appendix/prompts/understanding_q-cmdad.tex
\begin{figure*}[t]
\centering
\small
\noindent\begin{minipage}{1.0\linewidth}
\mdfsetup{%
middlelinewidth=1pt,
backgroundcolor=green!3,
innerleftmargin=0.5cm,
innerrightmargin=0.5cm,
roundcorner=15pt}
\begin{mdframed}
\vspace{0.2em}

You are a teacher who's job is to create questions out of a 1 line description of a clip from a movie to test narrative understanding of the students. The questions must ask about factual details related to the description.
The description is a 1 line summary, and the students are expected to answer the questions having watched the movie, without seeing the description.
\vspace{0.2cm}\\
\textbf{Examples}:\\
    A description such as "The shining spaceship lands on a strange planet covered in glowing blue plants and mist." can be converted into many questions such as , "Where does the spaceship land?" (Answer: On a strange planet), "What makes the planet unusual?" (Answer: The planet is covered in glowing blue plants and mist)\\
    A description such as "Mark waits alone by the lake after missing the last boat home." can be converted into many questions such as "Who is Mark with waiting by the lake?" (Answer: Mark is alone), "Why is Mark waiting by the lake?" (Answer: Mark missed the last boat home)

Remember that there may be many things happening in the clip from the movie, and the 1 line summary may choose to not highlight them. This may lead to ambiguous questions which should be avoided. \\

\textbf{Example of ambiguous question}:
    Question such as "What did the spaceship do?" is ambiguous given the description "The shining spaceship lands on a strange planet covered in glowing blue plants and mist.", because the spaceship might have done many things in the clip that were not described in the summary. The students will not know which action the question is asking for out of the many actions the spaceship performed in the clip.\\
    Question such as "What is the spaceship described as?" is ambiguous given the description "The shining spaceship lands on a strange planet covered in glowing blue plants and mist.", because the word "shining" used to describe the spaceship might exist only in the 1 line summary (to which the students don't have access to), and the spaceship "shining" may have not been the most prominent feature of the spaceship in the movie clip and so the students may consider the question ambiguous. \\

\textbf{Question and Answer Format}:\\

\textbf{Questions}: Limited to one or two lines, formulated to be insightful and not overtly indicative of the answer. Avoid using overly descriptive language that could hint at the correct answer.\\
\textbf{Answers}: Five options per question, formatted as "- A), - B), - C), - D), and - E)", concise and reflective of the question’s depth.\\
\textbf{Answer Key}: Specify the correct answer clearly with the formatting, "Correct Answer:", in the line following all the answer options.\\
\textbf{Rationale}: Write a rationale explaining the correctness of the "Answer Key" based on the scene’s description.\\

\textbf{Output format}:
Return json formatted text. Example:
{\fontsize{7}{8}\selectfont
\begin{verbatim}
[
    {
        "question": "question text1",
        "options": ["A) answer key 1", "B) answer key 2", "C) answer key 3", "D) answer key 4", "E) answer key 5"],
        "correct_answer": "E) answer key 5",
        "rationale": "<rationale>",
    },
    {
        "question": "question text1",
        "options": ["A) answer key 1", "B) answer key 2", "C) answer key 3", "D) answer key 4", "E) answer key 5"],
        "correct_answer": "A) answer key 1",
        "rationale": "<rationale>",
    }
]
\end{verbatim}
}

\textbf{Description}:
\{description\}

\end{mdframed}
\end{minipage}

\caption{Prompt used to generate \textcolor{ForestGreen}{\textbf{narrative understanding}} MCQAs for the CMD-AD dataset.}
\label{fig:understanding_q-cmdad}
\end{figure*}

%% file: appendix/prompts/understanding_q-mad.tex
\begin{figure*}[t]
\centering
\small
\noindent\begin{minipage}{1.0\linewidth}
\mdfsetup{%
middlelinewidth=1pt,
backgroundcolor=green!3,
innerleftmargin=0.5cm,
innerrightmargin=0.5cm,
roundcorner=15pt}
\begin{mdframed}
\vspace{0.2em}

You are a teacher who's job is to create questions from plot summary of a clip from a movie to test narrative understanding of the students. The questions must ask about factual details related to the plot.
The students are expected to answer the questions having watched the movie clip, without seeing the plot summary.\\


\textbf{Examples}:
    \begin{itemize}[noitemsep,topsep=0pt]
        \item A summary such as "The shining spaceship lands on a strange planet covered in glowing blue plants and mist." can be converted into many questions such as , "Where does the spaceship land?" (Answer: On a strange planet), "What makes the planet unusual?" (Answer: The planet is covered in glowing blue plants and mist)
        \item A summary such as "Mark waits alone by the lake after missing the last boat home." can be converted into many questions such as "Who is Mark with waiting by the lake?" (Answer: Mark is alone), "Why is Mark waiting by the lake?" (Answer: Mark missed the last boat home)
    \end{itemize}
Remember that there may be many things happening in the clip from the movie, and the summary may choose to not highlight them. This may lead to ambiguous questions which should be avoided. \\

\textbf{Example of ambiguous question}: 
    \begin{itemize}[noitemsep,topsep=0pt]
        \item Question such as "What did the spaceship do?" is ambiguous given the summary "The shining spaceship lands on a strange planet covered in glowing blue plants and mist.", because the spaceship might have done many things in the clip that were not described in the summary. The students will not know which action the question is asking for out of the many actions the spaceship performed in the clip.
        \item Question such as "What is the spaceship described as?" is ambiguous given the description "The shining spaceship lands on a strange planet covered in glowing blue plants and mist.", because the word "shining" used to describe the spaceship might exist only in the plot summary (to which the students don't have access to), and the spaceship "shining" may have not been the most prominent feature of the spaceship in the movie clip and so the students may consider the question ambiguous. \\\
    \end{itemize}

\textbf{Question and Answer Format}:\\

\textbf{Questions}: Limited to one or two lines, formulated to be insightful and not overtly indicative of the answer. Avoid using overly descriptive language that could hint at the correct answer.\\
\textbf{Answers}: Five options per question, formatted as "- A), - B), - C), - D), and - E)", concise and reflective of the question’s depth.\\
\textbf{Answer Key}: Specify the correct answer clearly with the formatting, "Correct Answer:", in the line following all the answer options.\\
\textbf{Rationale}: Write a rationale explaining the correctness of the "Answer Key" based on the scene’s description.\\

\textbf{Output format}:
Return json formatted text. Example:\\

{\fontsize{7}{8}\selectfont
\begin{verbatim}
[
    {
        "question": "question text1",
        "options": ["A) answer key 1", "B) answer key 2", "C) answer key 3", "D) answer key 4", "E) answer key 5"],
        "correct_answer": "E) answer key 5",
        "rationale": "<rationale>",
    },
    {
        "question": "question text1",
        "options": ["A) answer key 1", "B) answer key 2", "C) answer key 3", "D) answer key 4", "E) answer key 5"],
        "correct_answer": "A) answer key 1",
        "rationale": "<rationale>",
    }
]
\end{verbatim}
}

\textbf{Plot summary}
\{summary\}

\end{mdframed}
\end{minipage}

\caption{Prompt used to generate \textcolor{ForestGreen}{\textbf{narrative understanding}} MCQAs for the MAD-eval dataset.}
\label{fig:understanding_q_mad}
\end{figure*}

%% file: appendix/prompts/answering.tex
\begin{figure*}[t]
\centering
\small
\noindent\begin{minipage}{1.0\linewidth}
\mdfsetup{%
middlelinewidth=1pt,
backgroundcolor=green!3,
innerleftmargin=0.5cm,
innerrightmargin=0.5cm,
roundcorner=15pt}
\begin{mdframed}
\vspace{0.2em}
A series of questions and their options are given below. \\
\verb|{questions_with_choices}|\\

Provide 1 answer to each of the questions based on the following \verb|{context_type}|. \\
If \verb|{context_type}| are not available, then they will not be provided. Also come up with rationale for the answers, quoting the specific (one or more) \verb|{context_type}| used for 
answering the question.\\
If the rationale suggests that the question is answered by directly using \verb|{context_type}|, then the boolean variable \verb|{answered_from_var_name}| 
should be "True".\\
Otherwise, if the rationale suggests that the question is answered by not using \verb|{context_type}|, but by prior knowledge or by common sense, then the variable
\verb|{answered_from_var_name}| should be "False". \\
Always answer \verb|{answered_from_var_name}|  as either "True" with T upper case and "rue" lower case OR "False" with F upper case and "alse" lower case.\\

\verb|{context_type}|:
\verb|{context}| \\

\textbf{Instructions}
\begin{enumerate}[noitemsep,topsep=0pt]
    \item Every question has to be answered.
    \item There should be 1 and only 1 answer to each question. If no answer is known, take an educated guess. Do not answer the same question more than once.
    \item All questions should be answered independently, i.e., you may not use other questions and their options to answer any question.
    \item Answer only as in the output format provided:
\end{enumerate}

\vspace{0.2cm}

Output format (substitute the <...> with appropriate values):
\begin{verbatim}
[
    {
        "answer": "<answer>" ,
        "rationale": "<rationale>" ,
        "{answered_from_var_name}": "<{answered_from_var_name}>" ,
    },
    ...
]
\end{verbatim}

\vspace{0.2cm}

\end{mdframed}
\end{minipage}

\caption{Prompt used to answer questions from ADQA based on dialog and/or AD inputs.}
\label{fig:answering}
\end{figure*}